\documentclass{article}

\usepackage{setspace}
\usepackage{arxiv}
\usepackage{listings}
\usepackage{xcolor}
\usepackage[utf8]{inputenc} 
\usepackage[T1]{fontenc}    
\usepackage{newtxtext} 
\usepackage{newtxmath} 
\usepackage{tcolorbox}
\usepackage{float}  
\usepackage{hyperref}       
\usepackage{url}            
\usepackage{booktabs}       
\usepackage{amsfonts}       
\usepackage{nicefrac}       
\usepackage{microtype}      
\usepackage{cleveref}       
\usepackage{lipsum}         
\usepackage{graphicx}
\usepackage[numbers]{natbib}
\usepackage{doi}
\usepackage{amsmath}
\usepackage{geometry}
\usepackage{caption}

\title{Target Prompting for Information extraction with Vision Language Model}

\date{}

\lstdefinestyle{mystyle}{
    backgroundcolor=\color{white},   
    basicstyle=\ttfamily,            
    frame=single,                    
    breaklines=true,                 
    captionpos=b,                    
}

\newif\ifuniqueAffiliation
\uniqueAffiliationtrue

\ifuniqueAffiliation 
\author{ {\hspace{1mm}Dipankar Medhi} \\
}
\else
\fi

\renewcommand{\headeright}{}
\renewcommand{\undertitle}{}

\hypersetup{
pdftitle={Target Prompting for Information extraction with Vision Language Model},
pdfsubject={cs.AI},
pdfauthor={Dipankar Medhi},
pdfkeywords={Large Language Model, Prompt Engineering, Vision Language Model, Information Extraction, Information Retrieval},
}

\begin{document}
\maketitle

\begin{abstract}
The recent trend in the Large Vision and Language model has brought a new change in how information extraction systems are built. VLMs have set a new benchmark with their State-of-the-art techniques in understanding documents and building question-answering systems across various industries. They are significantly better at generating text from document images and providing accurate answers to questions. However, there are still some challenges in effectively utilizing these models to build a precise conversational system. General prompting techniques used with large language models are often not suitable for these specially designed vision language models. The output generated by such generic input prompts is ordinary and may contain information gaps when compared with the actual content of the document. To obtain more accurate and specific answers, a well-targeted prompt is required by the vision language model, along with the document image. In this paper, a technique is discussed called Target prompting, which focuses on explicitly targeting parts of document images and generating related answers from those specific regions only. The paper also covers the evaluation of response for each prompting technique using different user queries and input prompts. 
\end{abstract}


\section{Introduction}
\label{sec:introduction}

Understanding large documents and complex reports with the help of large language models \cite{kasneci2023chatgpt, chang2024survey} has revolutionized knowledge distribution and accessibility. With the rise of the large models in the industry \cite{rag_vs_finetuning_2024}, more and more fields are considering using these large models for their applications and incorporating them into their current or building new AI-powered systems, empowering streamlined usage by their users. 

The current large models are great at learning and storing factual information, but they have limited ability to access and manipulate this knowledge. To address these limitations, RAG pipelines were introduced \citep{lewis2020retrieval}. They combine a database (knowledge base) with the model \cite{gao2023retrieval}, expanding its memory and information pool without relying solely on the model's ability to learn new tasks as needed. These databases are specifically designed for RAG systems to quickly and accurately retrieve relevant information based on input queries. They are also known as Vector Databases, where the data is stored in the format of vector embeddings \cite{pan2024survey, han2023comprehensive, grohe2020word2vec}. 

\begin{figure}[h] 
    \centering 
    \includegraphics[width=0.85\textwidth]{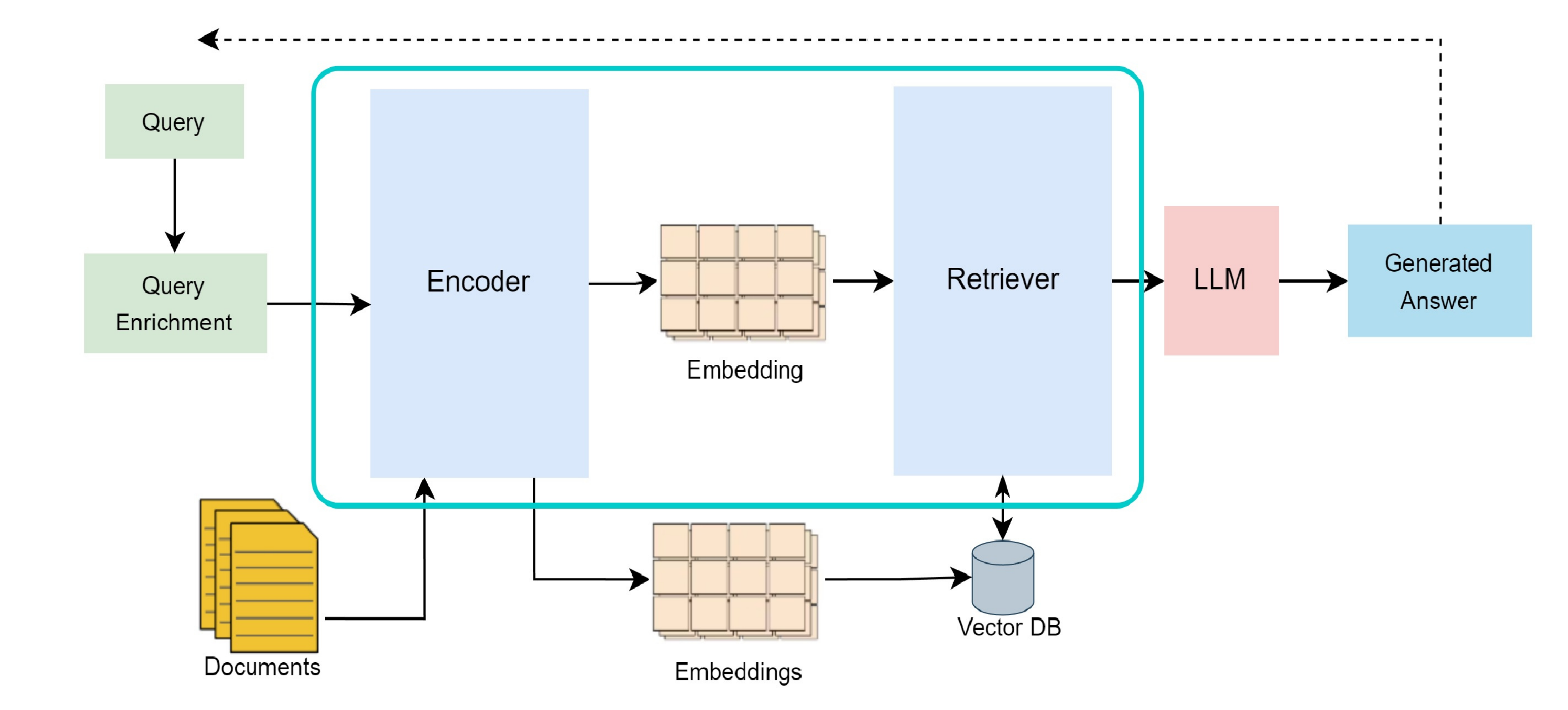} 
    \caption{\textbf{Overview of a RAG pipeline.} \emph{Encoder} converts text data to vector embeddings. \emph{Retriver} fetches relevant chunks from the vector store and feeds the retrieved information to \emph{LLM} to generate a response for the query.}
    \label{fig:ragpipeline} 
\end{figure}

The performance and accuracy of the RAG system heavily depend on how the data is processed and stored in the vector databases. For pipelines dealing with large documents and reports, it is crucial for the data or information extractor to perform well. Their ability to extract the correct information from the documents determines the quality of the responses by the generative model \cite[section 2.2]{rag_vs_finetuning_2024}. The better the quality of the data and the closer it is to the actual information in the documents, the better will be the model responses. Therefore, it is highly important to fine-tune the data extraction process. 

Document parsing tools like PDF and text parsers are suitable for digitally generated documents where the words are typed. However, the real challenge arises when the requirement is to extract information from image-based documents, where the text is not printed but present inside an image in a document. There are other alternatives such as Optical Character Recognition or OCR, that convert scanned or printed text images, and handwritten text, into digital text. Though this technology can solve basic text extraction tasks, it can face challenges while distinguishing similar types of characters like \emph{"0"} and \emph{"O"} \cite{patel2012optical}. This shows that there is a loss of information when OCR is used to extract information from complex document images. 

Understanding complex documents and complicated characters is a challenge. This is where multi-model pipelines show remarkable performance and do a comparatively reasonable job in information extraction \cite{zhang2021vinvl, zhang2024vision}. While Vision Language Models outperform other methods \cite{maeda2024vision}, challenges are still involved in accurately instructing the model to extract the required information from the documents. General prompting is beneficial when the goal is to get the overall gist of a document image, providing a well-structured summary of the document. However, the output response may lack complete information and insufficient details, even when asked to describe the document briefly. To solve this issue, this paper proposes a technique called "Target Prompting" that can help retrieve detailed information from a specific section of an image document, making the information extraction from documents, accurate and reasonable. 

\section{Method}
\label{sec:method}
The pipeline expects an image and a prompt for the model to generate text from the document image. A collection of open-source document images has been gathered and combined into a dataset for the experiment. Since the objective of this experiment is not to benchmark the method but to present a way of instructing the model for information extraction, only a handful of relevant images have been chosen for the study. 

\subsection{Model and Dataset}
\label{sec:modelanddataset}
\textbf{Phi-3-vision-instruct}. It is an open-sourced model by Microsoft with a parameter size of 4.2 Billion \cite{phi3_vision_2024}. It is a multimodal model that can process image and textual prompts, and generate textual outputs. This model has two primary components, an image encoder and a transformer decoder \cite[page 9]{phi3_vision_2024}.
The image encoder used in the Phi-3 Vision model is CLIP ViT-L/14 \cite{radford2021learning}. It is responsible for processing the visual information from the input image. The visual tokens extracted by the image encoder are combined with the text tokens in an interleaved manner, without any specific order.  
The transformer decoder is Phi-3-mini-128k-instruct \cite{phi3_vision_2024}. It is responsible for processing the combined visual and textual tokens to generate textual outputs based on the input image and text prompts.

\begin{figure}[h] 
    \centering 
     \includegraphics[width=0.8\textwidth]{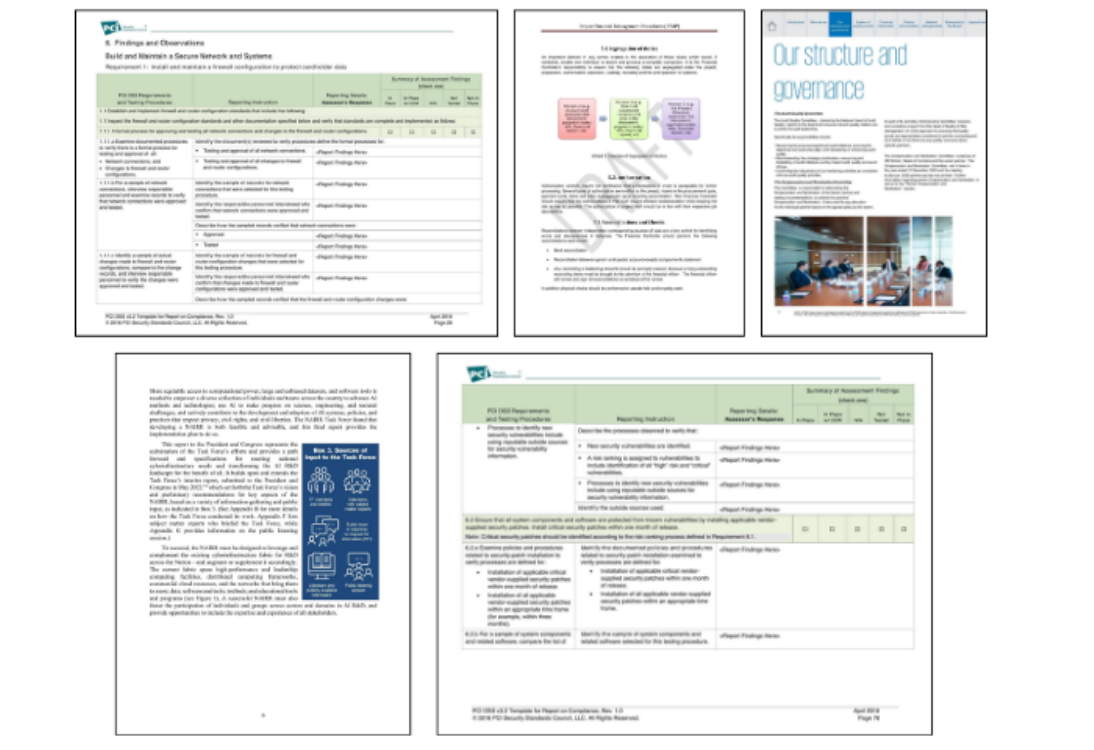}
     \caption{\textbf{Dataset sample document images}. Showcases a few sample document pages from the dataset.}
    \label{fig:dataset-document-sample}
\end{figure}

The dataset used for this experiment is \emph{vidore/syntheticDocQA\_government\_reports\_test\_captioning} from HuggingFace \cite{faysse2024colpaliefficientdocumentretrieval}. It has around 1900 rows of images, prompts and associated metadata. Out of all the available images, a subset of images is chosen based on \emph{Clarity}, \emph{Sharpness} and \emph{Resolution} and they are combined into a new dataset specifically for this experiment. The chosen images are handpicked, ensuring all of them are clean and not blurred, and have well-defined features. The selected images are paired with a specific prompt that targets a particular information from the document image that we intend to extract. See figure \ref{fig:dataset-document-sample} for the dataset sample.

\begin{table}[H]
  \centering
  \begin{tabular}{|c|c|}
    \hline
    \textbf{Image} & \textbf{Text Prompt} \\
    \hline
    Image 1 & Prompt 1\\
    \hline
    Image 2 & Prompt 2\\
    \hline
  \end{tabular}
  \vspace{4pt}
  \caption{Sample data subset}
  \label{tab:sampledataset}
\end{table}

\subsection{Information Extraction}
\label{sec:informationextraction}

\begin{figure}[H] 
    \centering 
    \includegraphics[width=0.7\textwidth]{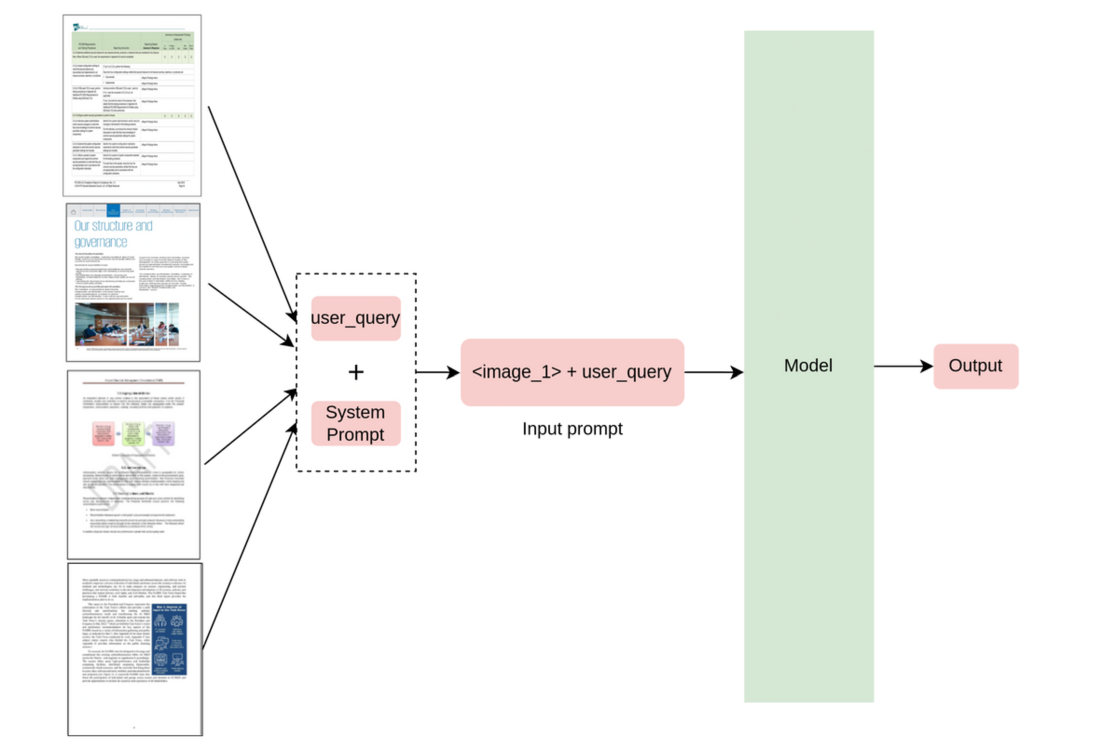}
     \caption{\textbf{Information extraction overview}. Model input includes \emph{document page image} and \emph{user query} with the system prompt "<image\_1>".}
    \label{fig:extractionprocess}
\end{figure}

The model and processor are initialized in an extraction function that takes in 2 arguments, \emph{image} and \emph{prompt}. All the image processing and tokenization are handled by the processor, followed by the answer generation by the model itself. The extraction process runs in a loop, iterating over all the selected images, feeding into the inference code along with the prompt, and generating results for each image one after another.  See figure \ref{fig:extractionprocess}. The input prompt is the aggregation of the \emph{user query} and the system prompt, \emph{<image\_1>}. This ensures that the prompt is in the correct format when fed into the information extraction function.

\section{Experiments}
\label{sec:experiments}

The experiment is primarily performed on the document images from the dataset, consisting of tabular pages and reports with infographics and charts. Since there are multiple entries with duplicate images, only a handful of them have been chosen for this experiment. The images are fed to the network with a user query and the system prompt, and the output is evaluated manually to rate its relevance. The model expects a system prompt \cite{phi3_vision_2024}, \emph{"<image\_1>"} as the image placeholder and following that an input query. See figure \ref{fig:extractionprocess} for an overview.

\textbf{General prompting}. Prompting is a method of guiding a large pre-trained model to tackle new tasks by providing specific instructions and task-specific hints \cite{sys_survey_vlm_2023}. For tasks like targeted question answering or data extraction from documents, the prompt needs to be tailored or adjusted to the specific section of the document from which the information is needed \cite{berrios2023towards}. 

When prompts are generalized, the resulting answers encompass the overall meaning of the image. If the aim is to grasp the general idea of a document or image, a prompt derived from a description query is adequate for the model to provide the image document's context \cite{li2022blip}. The model generates an answer that captures the whole meaning of the image document, yielding a more comprehensive response that covers every aspect of the image document.

\begin{figure}[H] 
    \centering 
    \includegraphics[width=0.9\textwidth]{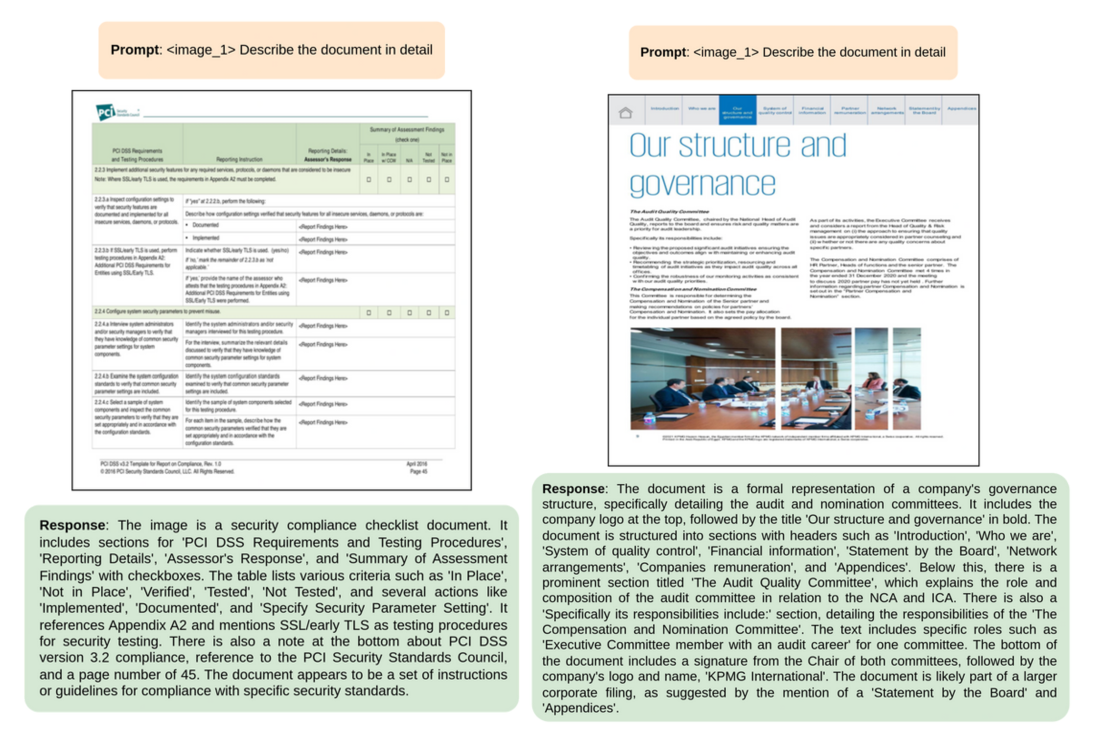} 
    \caption{\textbf{Overview of general prompt output}. The responses put all the details from the document images in a single chunk of text.} 
    \label{fig:generalprompting} 
\end{figure}

When the model is asked to provide a detailed description of an input image, the generated response contains a general overview of the image or document, spanning from top to bottom \cite{phi3_vision_2024}. However, When observed properly, it becomes clear that the model's response lacks specific details, indicating a difference between the generated content and the actual image document content. The information generated by the model is partially accurate, capturing only a limited amount of context from the original document. This pattern is observed across the majority of the documents utilized in the experiment. See figure \ref{fig:generalprompting} for sample examples.

\textbf{Target Prompting}. The problem of extracting targeted information from the image document using the model can be effectively achieved using target prompting. It has been observed that by explicitly instructing the model \cite{mirza2023tap} to describe a particular portion of the document image, it performs well compared to when it was asked a more general question \cite{arora2022ask}. This enables the model to perform tasks without covering everything from the input image object. By isolating regions of interest based on specific needs, it is possible to obtain precise answers more efficiently.

\begin{figure}[H] 
    \centering 
    \includegraphics[width=0.9\textwidth]{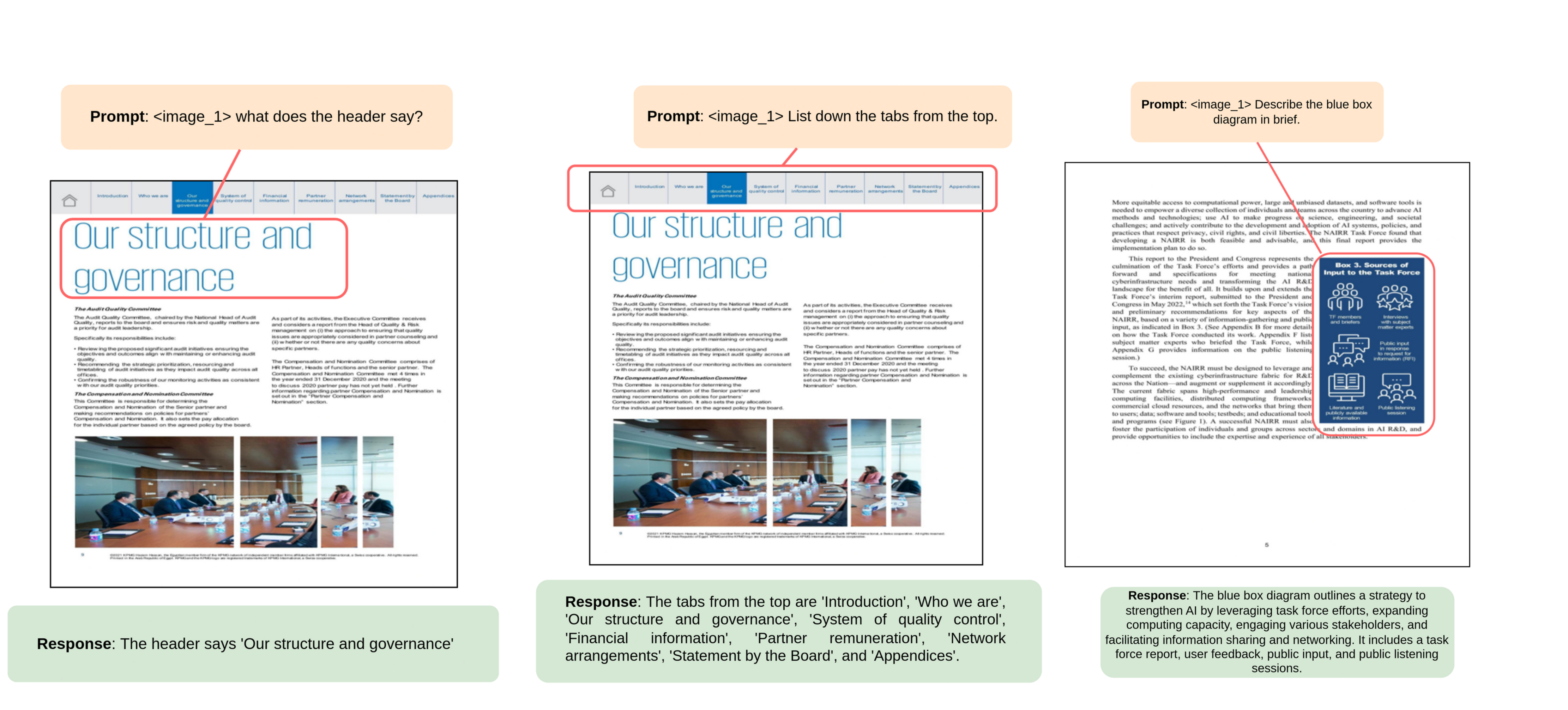} 
    \caption{\textbf{Target Prompting results}. The responses are more accurate and to the point. Gives more control over the model response.} 
    \label{fig:targetprompting} 
\end{figure}

Using target prompting, the questions are targeted to certain portions of the image document like diagrams, parts of the table, etc. \cite{mirza2023tap} This has made the answers more clear and precise, and closer to the expected response. The output does depend on the quality of the image -- higher-resolution images generally produce better results than lower-resolution images. 
Experiments with different types of image documents, including infographics, diagrams, tables, and long text paragraphs, showed that the models performed exceptionally well when asked targeted questions tailored to specific needs. See figure \ref{fig:targetprompting} for sample examples.

\subsection{Results}
\label{sec:results}

Focusing on a specific portion of an image and trying to extract information from that region alone can greatly improve the model performance and the final output accuracy. The target prompting technique has greatly improved the information extraction process by retrieving content from only those portions of the document images which has been pointed to in the prompt. This experiment is performed in an instance with 16 GB RAM and a Tesla T4 GPU.

The targeted prompts mostly include questions that are specific to a particular region or from a definite portion of the document image. See listing \ref{lst:targetpromptsamples}. This ensures that the generated output is free from any unwanted information or noise. The response is relevant to the question asked and only generates results related to the input query prompt. It enables a more rigid control over the type of responses to be generated from the documents. The same image can be fed multiple times to the extraction function to extract the desired information only, avoiding any irrelevant content and unwanted noise in the final answer. This has the potential to reduce the post-generation processes as the majority of the documents when processed with this technique will have only answers relevant to what is being asked in the prompt.

\lstset{style=mystyle}
\begin{lstlisting}[caption={Examples of information retrieval from document images using target prompting.}, label={lst:targetpromptsamples}]
Prompt: <image_1> What does the header say?
Response: The header says 'Our structure and governance'.

Prompt: <image_1> List down the tabs from the top.
Response: The tabs from the top are 'Introduction', 'Who we are', 'Our structure and governance', 'System of quality control', 'Financial information', 'Partner remuneration', 'Network arrangements', 'Statement by the Board', and 'Appendices'.

Prompt: <image_1> Describe the green portion of the table in detail.
Response: The green portion of the table contains the section titled 'Summary of Assessment Findings' with column headers '(check one)', 'In', 'In Place w/ CCW', 'N/A', 'Not Tested', 'Not in Place', and 'Reporting Details: Assessor's Response'. The corresponding entries are 'In', 'In Place w/ CCW', and 'N/A', followed by 'Not Tested', 'Not in Place', and 'Reporting Instruction'. Below this, there are detailed descriptions for various configurations and security features as per the PCI DSS requirements and additional policies.

Prompt: <image_1> Describe the blue box diagram in brief.
Response: The blue box diagram outlines a strategy to strengthen AI by leveraging task force efforts, expanding computing capacity, engaging various stakeholders, and facilitating information sharing and networking. It includes a task force report, user feedback, public input, and public listening sessions.
\end{lstlisting}

\section{Conclusion}
\label{sec:conclusion}

The paper proposed a specialized method to direct the vision language model, Phi-3-vision-instruct,  in generating specific responses related to the user input query. Although the results are not completely accurate, it performs significantly better when the task is to cover a certain portion of an image or document. 
There are more potential routes for improving the model responses and enhancing the accuracy of the proposed prompting technique, but that will require more testing and evaluation using a variety of datasets. In future work, the plan is to use an extended version of the dataset and conduct experiments to evaluate the performance and accuracy of the method on more complex documents.

\bibliographystyle{plain}
\bibliography{references}  






\end{document}